\documentclass{article}
\usepackage{spconf,amsmath,graphicx}
\usepackage{amsmath}
\usepackage{CJKutf8}
\usepackage{multirow}

\DeclareMathOperator*{\argmax}{argmax}

\title{Towards End-to-end Automatic Code-Switching Speech Recognition}
\name{Genta Indra Winata, Andrea Madotto, Chien-Sheng Wu, Pascale Fung\thanks{This work is partially funded by ITS/319/16FP of the Innovation Technology Commission, HKUST 16214415 \& 16248016 of Hong Kong Research Grants Council, and RDC 1718050-0 of EMOS.AI.}}
\address{
  Center for Artificial Intelligence Research (CAiRE)\\
  Department of Electronic and Computer Engineering\\
  Hong Kong University of Science and Technology, Clear Water Bay, Hong Kong\\ 
  \tt \{giwinata, amadotto, cwuak\}@connect.ust.hk, pascale@ece.ust.hk
}
\begin{document}
%\ninept
%
\maketitle
\begin{abstract}
\begin{CJK*}{UTF8}{gbsn}
Speech recognition in mixed language has difficulties to adapt end-to-end framework due to the lack of data and overlapping phone sets, for example in words such as "one" in English and "wàn" in Chinese. We propose a CTC-based end-to-end automatic speech recognition model for intra-sentential English-Mandarin code-switching. The model is trained by joint training on monolingual datasets, and fine-tuning with the mixed-language corpus. During the decoding process, we apply a beam search and combine CTC predictions and language model score. The proposed method is effective in leveraging monolingual corpus and detecting language transitions and it improves the CER by 5\%.
\end{CJK*}
\end{abstract}
\section{Introduction}
\label{sec:intro}
%Training speech recognition on mix languages with overlapping phonemes has been a huge challenge. 
Code-switching is the linguistic phenomenon of a person speaking or writing in one language and switches to another in the same sentence. It is very common among bilingual communities. With the advent of globalization, code-switching is becoming more common in predominantly monolingual societies as speakers use a second language in professional contexts. Speakers code-switch to empathize with each other, to express themselves better \cite{lowi2005codeswitching}, and very often are not fully aware of using mixed codes in their language \cite{shay2015switch}. 

Code-switching poses a significant challenge for automatic speech recognition (ASR) systems even as the latter reach higher and higher performance within the new paradigm of neural network speech recognition \cite{graves2006connectionist,amodei2016deep,Toshniwal2018MultilingualSR}. The main reason lies in the unpredictability of points of code-switching in an utterance. Since speech recognition needs to be accomplished in real-time in most applications, it is not feasible to carry out language identification at each time step before transcribing the speech into text. Moreover, language identification of a single speaker within the same utterance is challenging as the speaker can carry over their pronunciation habits from the primary language to the foreign language in context. In the statistical ASR framework, acoustic modeling, pronunciation modeling and language modeling of code-switching speech are carried out separately and assumed to be independent of each other. The hypothesis $\hat{Y}$ is normally calculated as the following:
\begin{equation}
    \hat{Y} = \argmax_Y{P(X|Y)P(Y)}
\end{equation}
where $X$ is the input signal, $Y$ is the target sequence, $P(X|Y)$ is the acoustic observations and $P(Y)$ is the language model. The acoustic model consists of multiple Hidden Markov phoneme models, is trained from both languages bilingually, where the pronunciation of phones from both languages are mapped to each other. Each shared or mapped phoneme model is then trained from speech samples in both languages. Variations of this type of bilingual acoustic models abound, but they are not difficult to train.

Another challenge is how to train the code-switching language model, $P(Y)$. Reliable statistical language models are derived from large amounts of text. In the case of code-switching speech, the data is often not enough to be generalizable. Previous work attempted to solve this problem by either generating more code-switching data by allowing code-switch at every point in the utterance, or at the phrasal boundaries \cite{li2012code} or even by phrasal alignment of parallel data according to linguistic constraints \cite{ Li2014CodeSL}. In each case, the code-switching points are not learned automatically. The generalizability of language models derived from this data is therefore doubtful. 

In this paper, we propose and investigate an entirely different approach of automatic code-switching speech recognition, using an end-to-end neural network framework to recognize speech from input spectrogram to output text, dispensing with the pipelined architecture of acoustic modeling followed by language modeling. We apply a joint-training in CTC-based speech recognition \cite{amodei2016deep} and finetune the model on code-switching dataset to learn the language transitions between them. We show that our proposed approach learns how to distinguish signals from different languages and achieves better results compared to the training only with a code-switching corpus. We compare the results by adding different amounts of the mixed-language corpus and test the effectiveness of the joint-training. In the decoding step, we rescore our generated sequences with an n-gram language model.

\section{Related Work}
\label{sec:related-work}
\noindent\textbf{Code-switching speech recognition: }The prior study on code-switching ASR is to incorporate a tri-phone HMMs as an acoustic model with an equivalence constraint in the language model \cite{li2012code}. \cite{adel2013recurrent} combined recurrent neural networks and factored language models to rescore the n-best hypothesis. \cite{Li2014CodeSL} proposed a lattice-based parsing to restrict the sequence paths to those permissible under the Functional Head Constraint. \cite{vu2012first} applied different phone merging approaches and combination with discriminative training. An extensive study on Hindi-English phone set sharing and gains improvement in WER
\cite{sivasankaran2018phone}. In another line of work, multi-task learning approaches in code-switching had been used for learning a shared representation of two or more different tasks on language modeling \cite{W18-3207} and acoustic model \cite{seltzer2013multi}.

\noindent\textbf{End-to-end approaches: } \cite{graves2006connectionist} presented the earliest implementation of CTC to end-to-end speech recognition.
A sequence-to-sequence model with attention \cite{luong2015effective} that learned to transcribe speech utterances to characters has been introduced by 
\cite{chan2016listen} as Listen-Attend-Spell (LAS). A multi-lingual approach was proposed by \cite{Toshniwal2018MultilingualSR} using Seq2Seq approach using a union of language-specific grapheme sets and train a grapheme-based sequence-to-sequence model jointly on data from all languages.~\cite{Hori2017AdvancesIJ} proposed to train a bilingual ASR for spontaneous Japanese and Chinese speech by using an end-to-end Seq2Seq model.

% the network is trained to perform both the primary classification task and one or more secondary tasks using a shared representation. The additional model parameters associated with the secondary tasks represent a minimal increase in the number of trained parameters and can be discarded at runtime.

% According to the linguists, this constraint is language independent \cite{belazi1994code}. \cite{winata2018code} proposed a multi-task framework to learn code-switching constraint from syntactic information by joint training. They showed the model learns the location of code-switching points.
% - Code-Switching ASR \\
% ~~ Traditional: AM + LM \\
% -- LM with multi-task learning  \\
% - End-to-end Speech Recognition \\
% ~~ CTC + LAS \\
% - Transfer Learning from monolingual

\section{Methodology}
\begin{CJK*}{UTF8}{gbsn}
In this section, we describe the joint training on our proposed end-to-end approach including the learning strategies. During the decoding stage, we add an external language model for rescoring. We denote our training sets, English monolingual dataset $\{(X_1^{en}, Y_1^{en}),..., (X_n^{en}, Y_n^{en})\}$ and Mandarin Chinese monolingual dataset $\{(X_1^{zh}, Y_1^{zh}),..., (X_n^{zh}, Y_n^{zh})\}$, and a code-switching dataset $\{(X_1^{cs}, Y_1^{cs}),..., (X_n^{cs}, Y_n^{cs})\}$. The labels $Y$ are graphemes and the character set is the concatenation of English and Simplified Chinese characters \textnormal{\tt \{a-z, space, apostrophe, 祥, 舌, ..., 底 \}}.
\end{CJK*}

\label{sec:methodology}
\begin{CJK*}{UTF8}{gbsn}
\subsection{Connectionist Temporal Classification Model}
A CTC network uses an error criterion that optimizes the prediction of transcription by aligning the input signals with the hypothesis. The loss function is defined as the negative log likelihood:
\begin{equation}
    L_{CTC} = - \textnormal{log } P(I|X) = - \textnormal{ln}\sum_{\pi\in B^{-1}(Y)}{P(\pi|X)}
\end{equation}
where we denote $\pi$ as the CTC path, $I$ as the transcription, and $B^{-1}(Y)$ as the mapping of all possible CTC paths $\pi$ resulting from $I$. Comparing to other end-to-end models such as LAS, CTC is more stable in training; thus it is easier to converge. Our model consists of a multi-layer Convolutional Neural Network \cite{lecun1998gradient} to generate a rich input representation, followed by a multi-layer Recurrent Neural Network (RNN) with Gated Recurrent Unit (GRU) \cite{cho2014learning} to learn the temporal information of the audio frame sequences. We are using the CTC-based architecture similar to \cite{amodei2016deep}. The input of the network is a sequence of log spectrograms and followed by a normalization step to regularize the parameters and avoid internal covariate shift.  The CTC probabilities are used to find a better alignment between hypothesis and the input signals. To scale up the training process, batch normalization is employed to the recurrent layers to reduce the batch training time. The spectrograms are passed to a convolutional neural network (CNN) to encode the speech representation. The frame-wise posterior distribution $P(Y|X)$ is conditioned on the input $X$ and calculated by applying a fully-connected layer and a softmax function.
\begin{equation}
    P(Y|X) = \textnormal{Softmax}(\textnormal{Linear}(h))
\end{equation}
where $h$ is the hidden state from the GRU. Next, it passes to a multi-layer bidirectional GRU.
\begin{figure}[!t]
  \centering
  \includegraphics[width=0.92\linewidth]{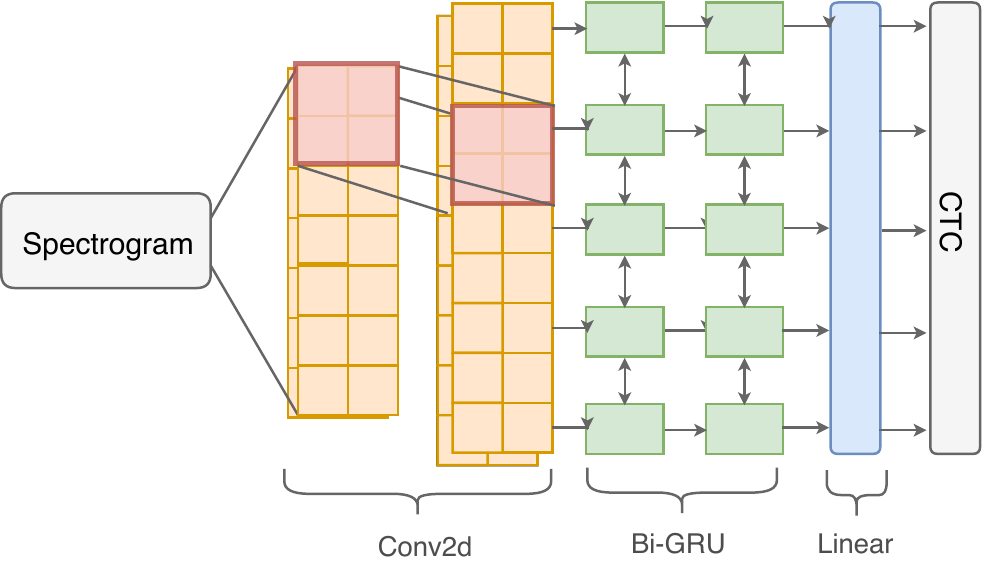}
  \caption{Connectionist Temporal Classification Model}
  \label{fig:ctc}
\end{figure}
\subsection{Joint Training}
We start the training as a joint training using monolingual dataset as a pretraining to learn the individual language. However, the CTC-based model can easily suffer catastrophic forgetting, where the model is not capable of remembering two distant languages such as English and Mandarin Chinese in separated supervision. It tends to keep one language and forget the other. Thus, we propose to train iteratively between two datasets; taking English and Chinese speeches in the batch. Thus, the model learns how to differentiate the characters. After the pretraining, we tune the model with code-switch data.
\end{CJK*}

\subsection{Language Modeling}
To improve the quality of the decoded sequence, we train a 5-gram language model with Kneser Ney smoothing \cite{heafield2013scalable} on our code-switching training data using KenLM\footnote{The code can be found at https://github.com/kpu/kenlm}. We use a prefix beam search with a beam width of $w$. We rescore the probability of the sequence $p_lm{(Y)}$  \cite{amodei2016deep} and find the maximum score of $Q(Y)$:
\begin{equation}
    Q(Y) = \textnormal{log}(P_{ctc}(Y|X)) + \alpha \textnormal{ log}(p_{lm}(Y)) + \beta \textnormal{ } wc(Y)
\end{equation}
where $wc(Y)$ is the word count of sequence $Y$, $\alpha$ controls the contribution of language model and $\beta$ controls the number of word to be generated. In the beam search process, the decoder computes a score of each partial hypothesis and interpolates the result with the probability from the language model.

\section{Experiment}
In this section, we describe our datasets and code-switching ASR experiments with our
proposed end-to-end system.

\begin{table}[!t]
\centering
\caption{Data Statistics of SEAME Phase II \cite{W18-3207}.}
\label{data-statistics-phase-2}
\begin{tabular}{@{}rccc@{}}
\hline
\multicolumn{1}{|l|}{} & \multicolumn{1}{c|}{\textbf{Train}} & \multicolumn{1}{c|}{\textbf{Dev}} & \multicolumn{1}{c|}{\textbf{Test}} \\ \hline
\multicolumn{1}{|r|}{\# Speakers}               & \multicolumn{1}{c|}{138} & \multicolumn{1}{c|}{8} & \multicolumn{1}{c|}{8} \\ \hline
\multicolumn{1}{|r|}{\# Duration (hr)} & \multicolumn{1}{c|}{100.58} & \multicolumn{1}{c|}{5.56} & \multicolumn{1}{c|}{5.25} \\ \hline
\multicolumn{1}{|r|}{\# Utterances} & \multicolumn{1}{c|}{78,815} & \multicolumn{1}{c|}{4,764} & \multicolumn{1}{c|}{3,933} \\ \hline
\multicolumn{1}{|r|}{\# Tokens} & \multicolumn{1}{c|}{1.2M} & \multicolumn{1}{c|}{65K} & \multicolumn{1}{c|}{60K} \\ \hline
\multicolumn{1}{|r|}{\begin{tabular}[c]{@{}r@{}}\# Tokens\\ Preprocessed\end{tabular}} & \multicolumn{1}{c|}{978K} & \multicolumn{1}{c|}{53K} & \multicolumn{1}{c|}{48K} \\ \hline
\multicolumn{1}{|r|}{Avg. segment} & \multicolumn{1}{c|}{4.21} & \multicolumn{1}{c|}{3.59} & \multicolumn{1}{c|}{3.99}
\\ \hline
\multicolumn{1}{|r|}{Avg. switches}                                                 & \multicolumn{1}{c|}{2.94}                    & \multicolumn{1}{c|}{3.12}                  & \multicolumn{1}{c|}{3.07}               \\ \hline
\end{tabular}
\end{table}

\begin{table}[!t]
\centering
\small
\caption{Data Statistics of Common Voice and HKUST}
\begin{tabular}{|l|l|c|c|}
\hline
\multicolumn{2}{|c|}{\textbf{Dataset}} & \textbf{\# Duration (hr)} & \textbf{\# Samples} \\ \hline
\multirow{3}{*}{Common Voice}& \textbf{Train} & 241.21 & 195,372 \\
& \textbf{Dev} & 4.99 & 4,065 \\
& \textbf{Test} & 4.94 & 3,986 \\ \hline
\multirow{2}{*}{HKUST} & \textbf{Train} & 168.88 & 873 \\
& \textbf{Dev} & 4.81 & 24 \\ \hline
\end{tabular}
\end{table}

% \begin{table}[!t]
% \centering
% \caption{N-gram language model performance (Perplexity)}
% \label{lm}
% \begin{tabular}{l|l|l}
% \hline
% \textbf{Model} & \multicolumn{1}{c|}{\textbf{Dev}} & \multicolumn{1}{c}{\textbf{Test}} \\ \hline \hline
% 3-gram         & 219.62                            & 215.35                             \\
% 4-gram         & 218.83                            & 214.13                             \\
% 5-gram         & 218.41                            & 213.75                             \\ \hline
% \end{tabular}
% \end{table}

\subsection{Corpus}
We use speech data from SEAME Phase II (South East Asia Mandarin-English), a conversational
Mandarin-English code-switching speech corpus consists of spontaneously spoken interviews and conversations \cite{SEAME2015}. We tokenize words using Stanford NLP toolkit~\cite{manning-EtAl:2014:P14-5} and follow the same preprocessing step as \cite{W18-3207}. For the monolingual datasets, we use HKUST \cite{liu2006hkust}, a spontaneous Mandarin Chinese telephone speech recordings and Common Voice, an open accented English dataset collected by Mozilla\footnote{The dataset is available at https://voice.mozilla.org/}. Table \ref{data-statistics-phase-2} shows the statistics of SEAME Phase II dataset and Table \ref{data-statistics-cv-hkust} shows the statistics of Common Voice and HKUST datasets.

\begin{table}[!t]
\centering
\caption{Character Error Rate (CER \%) for single dataset training, joint training, and rescoring with LM on SEAME Phase II.}
\label{results}
\begin{tabular}{l|c|c}
\hline
\textbf{Model} & \textbf{Dev} & \textbf{Test} \\ \hline
% \multirow{3}{*}{CTC 2-layer} & baseline (beam)& 37.70 & 31.80 \\
% \multirow{3}{*}{CTC 2-layer} (greedy) & baseline & 38.93 & 33.20 \\
% & joint training & 63.6 & 62.447 \\
% & joint training (greedy) & 63.44 & 62.14 \\
% & + fine tuning (beam) & 36.24 & 29.67 \\
% & + fine tuning (greedy)& 36.99 & 30.75 \\
% & + LM (beam $\alpha$=0.5) & 35.93 & 28.93 \\
\hline
% & + LM (greedy) & 35.93 & 28.93 \\
% \hline
% \multirow{3}{*}{CTC 4-layer} & baseline (greedy:7) & 36.84 & 30.83 \\
\textbf{SEAME Phase II} & & \\
- training (10\% data, 10.13 hr) & 49.81 & 46.23 \\
- training (50\% data, 50.41 hr) & 38.08 & 32.10 \\
% - training (75\%) &  33.34 & 27.55\\
- training (100\% data, 100.58 hr) & 36.18 & 29.82 \\
\textnormal{\enspace\enspace} + LM (5-gram, $\alpha$ = 0.2) &  35.77 & 29.14 \\ \hline
% (en) CommonVoice & 74.56 & 76.36 \\
% (zh) HKUST & 66.30 & 64.30 \\
% (en-zh) joint training & 64.00 & 61.58 \\
\textbf{Joint training} & & \\
- fine tuning (10\% data, 10.13 hr) & 38.44 & 43.86\\
- fine tuning (50\% data, 50.41 hr) & 34.24 & 27.97 \\
% - fine tuning (75\%) &  31.33 & 25.17 \\
- fine tuning (100\% data, 100.58 hr) & 32.06 & 25.54 \\
\textnormal{\enspace\enspace} + LM (5-gram, $\alpha$ = 0.2) &  \textbf{31.35} & \textbf{24.61} \\ \hline

% + LM ($\alpha$=0.25) &  \textbf{31.39} & \textbf{24.54} \\
% + LM ($\alpha$=0.5) & 31.64 & 24.69 \\ 
% + LM ($\alpha$=0.75) & 32.91 & 25.81 \\ \hline

% & joint training (greedy:7) & 64.00 & 61.58 \\
% & joint training (beam:7) & 66.521 & 64.756
% & joint training & 66.30 & 65.40 \\
% & + fine tuning (greedy) & 32.94 & 26.57 \\
% \textbf{LM} & & \\
% \textnormal{-} 3-gram & 219.62 & 215.35 \\
% \textnormal{-} 4-gram & 218.83 & 214.13 \\
% \textnormal{-} 5-gram & 218.41 & 213.75 \\ \hline
\end{tabular}
\end{table}

\begin{CJK*}{UTF8}{gbsn}
\begin{table*}[!t]
\centering
\caption{Generated Sequences from CTC 4-layer GRU }
\label{data-statistics-cv-hkust}
\label{generated-sentences}
\begin{tabular}{|l|p{11cm}|c|}
\hline
\textbf{Model} & \textbf{Generated Sequence} & \multicolumn{1}{l|}{\textbf{CER \%}} \\ \hline
reference & 因为\enspace我\enspace的 \enspace friend\enspace会\enspace很\enspace shy\enspace我\enspace的\enspace friend\enspace他\enspace这\enspace not really\enspace很\enspace shy\enspace他们\enspace就是\enspace要\enspace人家\enspace陪\enspace他们\enspace唱\enspace因为\enspace他们\enspace觉得\enspace一个人\enspace唱\enspace很\enspace sian & - \\
baseline & 为 \enspace 我的 \enspace \textnormal{friend wa} \enspace 很 \enspace s \enspace 我们 \enspace 在\enspace friend \enspace 他 \enspace 这\enspace not ra \enspace 能 \enspace s \enspace 他 \enspace 就 \enspace 要 \enspace 人家 \enspace pa \enspace 他们 \enspace 唱 \enspace 因为 \enspace 他们 \enspace 觉得 \enspace 一个 \enspace 人 \enspace 唱 \enspace 很 \enspace 闲 \enspace & 23.65\% \\
+ fine-tuning & 因为\enspace我的\enspace friend \enspace会\enspace 很\enspace shy \enspace 我\enspace friend \enspace他\enspace做\enspace not re \enspace很\enspace帅\enspace他们\enspace就\enspace要\enspace人家\enspace陪\enspace他们\enspace唱\enspace因为\enspace他们\enspace觉得\enspace一\enspace个人\enspace唱\enspace很\enspace鲜\enspace & 15.05\% \\
+ LM & 因为 \enspace我\enspace的\enspace friend\enspace会\enspace很\enspace shy\enspace 我\enspace friend\enspace 他\enspace not real s \enspace他们\enspace就\enspace要\enspace人家\enspace陪\enspace他们\enspace唱\enspace因为\enspace他们\enspace觉得\enspace一\enspace个\enspace人\enspace唱\enspace很\enspace s \enspace & 11.82\% \\ \hline
reference & then\enspace你\enspace做\enspace什么\enspace before what kind of job & - \\ 
baseline & 你\enspace做\enspace什么\enspace可是\enspace what\enspace他\enspace要\enspace dro & 48.57\%  \\
+ fine-tuning & 因\enspace你\enspace做\enspace什\enspace么\enspace for what kid of job & 22.85\% \\
+ LM & 你\enspace做\enspace什\enspace么\enspace for what kind of job & 20.00\% \\ \hline
\end{tabular}
\end{table*}
\end{CJK*}

\subsection{Experimental Setup}
We convert the inputs into normalized frame-wise spectrograms. We take 20 ms width with a stride of 20 ms. The audio is down-sampled to a single channel with a sample rate of 8 kHz. The CNN encoder is described as the following:
\begin{align}
    &\textnormal{Conv2d}(\textnormal{in}=1, \textnormal{out}=32, \textnormal{filter}=41\times11)\\
    &\textnormal{Hardtanh(BatchNorm2d}(32))\\
    &\textnormal{Conv2d}(\textnormal{in}=32, \textnormal{out}=32, \textnormal{filter}=21\times11)\\
    &\textnormal{Hardtanh(BatchNorm2d}(32))
\end{align}
It is followed by a 4-layer bidirectional GRU with a hidden size of 400 and a fully connected layer with a hidden size of 400 is inserted afterward. In the joint-training, we combine both monolingual datasets and sorted by their audio length, and groups them in buckets. We shuffle the buckets and take 20 samples in a batch. Then, we take every batch for the training.

We start our training with different initial learning rates  $\{1e-4, 3e-4\}$ and optimize our model by using Stochastic Gradient Descent (SGD) with momentum and Nesterov accelerated gradient \cite{nesterov1983method}. In the sequence decoding stage, we take to run a prefix beam-search a beam size of 100 to find the best sequence. The best hyperparameters for rescoring are $\alpha = 0.2$ and $\beta=1$. We first perform our fine-tuning experiment with smaller training sets to calculate the effectiveness of joint training. We take 10\% (10.13 hr) and 50\% (50.41 hr) from code-switching dataset. We also train our language model with 3-gram, 4-gram, and 5-gram models.

\section{Results}
\label{sec:results}

Table \ref{results} demonstrates speech recognition and language modeling results. The joint training with fine tuning improves the results 5\% CER compared to training only with a code-switching corpus. Some generated characters are separated with excessive spaces. The joint training captures the sounds from monolingual corpus and transfer learns the information during the fine-tuning step. The baseline is not able to capture the word transition between Chinese and English. The joint training captures the sounds from the monolingual corpus and transfers learning to the code-switching sequences.

\noindent\textbf{Joint training + Fine tuning: } Joint training helps the model to learn different sounds the perspective of single language. It is also an excellent way to initialization of our model. However, from the training, we still suffer an issue, where it does not keep the information of both languages, and we need to solve this issue in the future work. As shown in Table \ref{generated-sentences}, we can see that the results are getting better, it can generate a  better mapping of similar sounds in English and Chinese to the corresponding characters. We test our model with smaller training data. According to our observation, by fine-tuning with only 50\% code-switching data, we can achieve a comparable result to the whole training only with code-switching dataset.

\noindent\textbf{Applying language model: } The external language model constraints the decoder to generate more grammatical sentences. In general, language model effects positively to the decoder and achieves an additional 1\% reduction by adding a 5-gram language model. From Table \ref{generated-sentences}, it clearly shows that some misspelled words in the baseline are fixed.

\noindent\textbf{Intra-sentential code-switching: }  The model we trained can predict some English words correctly between Chinese words testing on code-switching dataset. It can still predict the code-switching points, unlike the work by \cite{Toshniwal2018MultilingualSR}. One of the possible reason is our CTC model is not constrained by the language information like the Seq2Seq-based model with language identifiers. In spite of that, the model is still predicting words with similar sound in the code-switching points such as ``before" and ``for". 

\section{Conclusion}
\label{sec:conclusion}
We propose a new direction on automatic code-switching speech recognition by applying end-to-end approach. Our training method can be adapted to any languages pair. We evaluate our model on English-Mandarin corpus and achieve a significant gain through a combination of joint training and fine-tuning. It can handle code-switching transitions and recognize both English and Chinese characters. The rescoring using an external language model improves the decoding result and fixes the spelling mistakes. Our proposed model achieves a 5\% reduction in CER and the joint-training procedure allows the model to learn distant languages. For future work, we are going to mitigate further the catastrophic forgetting in our joint training network, which degrades the performance of our bilingual model.

% \section{REFERENCES}
% \label{sec:refs}

% List and number all bibliographical references at the end of the
% paper. The references can be numbered in alphabetic order or in
% order of appearance in the document. When referring to them in
% the text, type the corresponding reference number in square
% brackets as shown at the end of this sentence \cite{C2}. An
% additional final page (the fifth page, in most cases) is
% allowed, but must contain only references to the prior
% literature.

\bibliographystyle{IEEEbib}
\bibliography{strings,refs}

\begin{thebibliography}{10}

\bibitem{lowi2005codeswitching}
Rosamina Lowi,
\newblock ``Codeswitching: An examination of naturally occurring
  conversation,''
\newblock in {\em Proceedings of the 4th International Symposium on
  Bilingualism}. Cascadilla Press, pp Somerville, MA, 2005, pp. 1393--1406.

\bibitem{shay2015switch}
Orit Shay,
\newblock ``To switch or not to switch: Code-switching in a multilingual
  country,''
\newblock {\em Procedia-Social and Behavioral Sciences}, vol. 209, pp.
  462--469, 2015.

\bibitem{graves2006connectionist}
Alex Graves, Santiago Fern{\'a}ndez, Faustino Gomez, and J{\"u}rgen
  Schmidhuber,
\newblock ``Connectionist temporal classification: labelling unsegmented
  sequence data with recurrent neural networks,''
\newblock in {\em Proceedings of the 23rd international conference on Machine
  learning}. ACM, 2006, pp. 369--376.

\bibitem{amodei2016deep}
Dario Amodei, Sundaram Ananthanarayanan, Rishita Anubhai, Jingliang Bai, Eric
  Battenberg, Carl Case, Jared Casper, Bryan Catanzaro, Qiang Cheng, Guoliang
  Chen, et~al.,
\newblock ``Deep speech 2: End-to-end speech recognition in english and
  mandarin,''
\newblock in {\em International Conference on Machine Learning}, 2016, pp.
  173--182.

\bibitem{Toshniwal2018MultilingualSR}
Shubham Toshniwal, Tara~N. Sainath, Ron~J. Weiss, Bo~Li, Pedro~J. Moreno,
  Eugene Weinstein, and Kanishka Rao,
\newblock ``Multilingual speech recognition with a single end-to-end model,''
\newblock {\em 2018 IEEE International Conference on Acoustics, Speech and
  Signal Processing (ICASSP)}, pp. 4904--4908, 2018.

\bibitem{li2012code}
Ying Li and Pascale Fung,
\newblock ``Code-switch language model with inversion constraints for mixed
  language speech recognition,''
\newblock {\em Proceedings of COLING 2012}, pp. 1671--1680, 2012.

\bibitem{Li2014CodeSL}
Ying Li and Pascale Fung,
\newblock ``Code switch language modeling with functional head constraint,''
\newblock {\em 2014 IEEE International Conference on Acoustics, Speech and
  Signal Processing (ICASSP)}, pp. 4913--4917, 2014.

\bibitem{adel2013recurrent}
Heike Adel, Ngoc~Thang Vu, Franziska Kraus, Tim Schlippe, Haizhou Li, and Tanja
  Schultz,
\newblock ``Recurrent neural network language modeling for code switching
  conversational speech,''
\newblock in {\em Acoustics, Speech and Signal Processing (ICASSP), 2013 IEEE
  International Conference on}. IEEE, 2013, pp. 8411--8415.

\bibitem{vu2012first}
Ngoc~Thang Vu, Dau-Cheng Lyu, Jochen Weiner, Dominic Telaar, Tim Schlippe,
  Fabian Blaicher, Eng-Siong Chng, Tanja Schultz, and Haizhou Li,
\newblock ``A first speech recognition system for mandarin-english code-switch
  conversational speech,''
\newblock in {\em Acoustics, Speech and Signal Processing (ICASSP), 2012 IEEE
  International Conference on}. IEEE, 2012, pp. 4889--4892.

\bibitem{sivasankaran2018phone}
Sunit Sivasankaran, Brij Mohan~Lal Srivastava, Sunayana Sitaram, Kalika Bali,
  and Monojit Choudhury,
\newblock ``Phone merging for code-switched speech recognition,''
\newblock in {\em Proceedings of the Third Workshop on Computational Approaches
  to Linguistic Code-Switching}, 2018, pp. 11--19.

\bibitem{W18-3207}
Genta~Indra Winata, Andrea Madotto, Chien-Sheng Wu, and Pascale Fung,
\newblock ``Code-switching language modeling using syntax-aware multi-task
  learning,''
\newblock in {\em Proceedings of the Third Workshop on Computational Approaches
  to Linguistic Code-Switching}. 2018, pp. 62--67, Association for
  Computational Linguistics.

\bibitem{seltzer2013multi}
Michael~L Seltzer and Jasha Droppo,
\newblock ``Multi-task learning in deep neural networks for improved phoneme
  recognition,''
\newblock in {\em Acoustics, Speech and Signal Processing (ICASSP), 2013 IEEE
  International Conference on}. IEEE, 2013, pp. 6965--6969.

\bibitem{luong2015effective}
Thang Luong, Hieu Pham, and Christopher~D Manning,
\newblock ``Effective approaches to attention-based neural machine
  translation,''
\newblock in {\em Proceedings of the 2015 Conference on Empirical Methods in
  Natural Language Processing}, 2015, pp. 1412--1421.

\bibitem{chan2016listen}
William Chan, Navdeep Jaitly, Quoc Le, and Oriol Vinyals,
\newblock ``Listen, attend and spell: A neural network for large vocabulary
  conversational speech recognition,''
\newblock in {\em Acoustics, Speech and Signal Processing (ICASSP), 2016 IEEE
  International Conference on}. IEEE, 2016, pp. 4960--4964.

\bibitem{Hori2017AdvancesIJ}
Takaaki Hori, Shinji Watanabe, Yu~Zhang, and William Chan,
\newblock ``Advances in joint ctc-attention based end-to-end speech recognition
  with a deep cnn encoder and rnn-lm,''
\newblock in {\em INTERSPEECH}, 2017.

\bibitem{lecun1998gradient}
Yann LeCun, L{\'e}on Bottou, Yoshua Bengio, and Patrick Haffner,
\newblock ``Gradient-based learning applied to document recognition,''
\newblock {\em Proceedings of the IEEE}, vol. 86, no. 11, pp. 2278--2324, 1998.

\bibitem{cho2014learning}
Kyunghyun Cho, Bart van Merrienboer, Caglar Gulcehre, Dzmitry Bahdanau, Fethi
  Bougares, Holger Schwenk, and Yoshua Bengio,
\newblock ``Learning phrase representations using rnn encoder--decoder for
  statistical machine translation,''
\newblock in {\em Proceedings of the 2014 Conference on Empirical Methods in
  Natural Language Processing (EMNLP)}, 2014, pp. 1724--1734.

\bibitem{heafield2013scalable}
Kenneth Heafield, Ivan Pouzyrevsky, Jonathan~H Clark, and Philipp Koehn,
\newblock ``Scalable modified kneser-ney language model estimation,''
\newblock in {\em Proceedings of the 51st Annual Meeting of the Association for
  Computational Linguistics (Volume 2: Short Papers)}, 2013, vol.~2, pp.
  690--696.

\bibitem{SEAME2015}
Universiti Sains~Malaysia Nanyang Technological~University,
\newblock ``Mandarin-english code-switching in south-east asia ldc2015s04. web
  download. philadelphia: Linguistic data consortium,'' 2015.

\bibitem{manning-EtAl:2014:P14-5}
Christopher~D. Manning, Mihai Surdeanu, John Bauer, Jenny Finkel, Steven~J.
  Bethard, and David McClosky,
\newblock ``The {Stanford} {CoreNLP} natural language processing toolkit,''
\newblock in {\em Association for Computational Linguistics (ACL) System
  Demonstrations}, 2014, pp. 55--60.

\bibitem{liu2006hkust}
Yi~Liu, Pascale Fung, Yongsheng Yang, Christopher Cieri, Shudong Huang, and
  David Graff,
\newblock ``Hkust/mts: A very large scale mandarin telephone speech corpus,''
\newblock in {\em Chinese Spoken Language Processing}, pp. 724--735. Springer,
  2006.

\bibitem{nesterov1983method}
Yurii~E Nesterov,
\newblock ``A method for solving the convex programming problem with
  convergence rate o (1/k\^{} 2),''
\newblock in {\em Dokl. Akad. Nauk SSSR}, 1983, vol. 269, pp. 543--547.

\end{thebibliography}

\end{document}